Assessing effect sizes, variability, and power in the on-line study of language production

Audrey Bürki[1,2] & Shravan Vasishth[1]

[1] Cognitive Science, University of Potsdam

[2] Linguistic Research Infrastructure, University of Zürich

Author note

Correspondence concerning this article should be addressed to Audrey Bürki, Karl-Liebknecht-Straße 24-25, 14476 Potsdam, Germany, 00493319772986. E-mail: buerki@uni-potsdam.de

Abstract

With the pandemic, many experimental psychologists/linguists have started to collect data over the internet (hereafter "on-line data"). The feasibility of such experiments and the sample sizes required to achieve sufficient statistical power in future experiments have to be assessed. This in turn requires information on effect sizes and variability. In a series of analyses, we compare response time data obtained in the same word production experiment conducted in the lab and on-line. These analyses allow us to determine whether the two settings differ in effect sizes, in the consistency of responses over the course of the experiment, in the variability of average response times across participants, in the magnitude of effect sizes across participants, or in the amount of unexplained variability. We assess the impact of these differences on the power of the design in a series of simulations. Our findings temper the enthusiasm raised by previous studies and suggest that on-line production studies might be feasible but at a non-negligible cost. The sample sizes required to achieve sufficient power in on-line language production studies come with a non-negligible increase in the amount of manual labour.

Assessing effect sizes, variability, and power in the on-line study of language production

## Introduction

The pandemic has accelerated a small technical revolution in experimental psychology. Many labs have had no choice but to turn to the internet to collect their data. Some fields have been more reluctant to go for it, the worry being that the data may be too noisy to allow effects to be detected. For instance, this is the case in psycholinguistics where experiments often rely on response times and effects can be rather small. Several authors have expressed the worry that the increased level of noise in data collected over the internet could prevent these effects from being accurately estimated (e.g., Kim, Gabriel, & Gygax, 2019). If it turns out that data collected over the internet (hereafter, "on-line data") allow the detection of psycholinguistic effects in a reliable way, the advantages would be numerous. Researchers would be able to access more diverse types of languages, contributing to the generalization of existing models to other languages or leading to refinements of these models in the light of cross-linguistic differences. Most studies in psycholinguistics indeed deal with a very small set of languages (Blasi, Henrich, Adamou, Kemmerer, & Majid, 2022; but see Jäger & Norcliffe, 2009; Tsegaye, Mous, & Schiller, 2014). Furthermore, on-line data could also be informative about how well current models fare in accounting for psycholinguistic processes beyond the population of linguistics and psychology students. Perhaps most importantly, however, the opportunity to collect data over the internet could allow researchers to increase their samples of participants and finally start collecting data with sufficient power. Most studies in the field are underpowered (Brysbaert, 2019; Bürki, Elbuy, Madec, & Vasishth, 2020; Vasishth, Mertzen, Jäger, & Gelman, 2018). As a consequence, there is a high probability of type II error (i.e., not finding an effect that exists), Type S, and Type M errors (i.e., reporting significant effects with respectively the wrong sign or an exaggerated effect size, see Gelman & Carlin, 2014). Given that participant recruitment over the internet is presumably easier and faster, the same experiment could also be more easily replicated before publication. Before we get too enthusiastic about on-line data collection, however, we need to assess the quality of these data and whether they are indeed suitable for (at least some of) our experimental designs. This in turn requires information on true effect sizes in on-line settings and on the amount and sources of variability in the data. The aim of the present study is to provide a direct comparison of effect sizes and variability for two datasets collected with the same experiment but in different settings. The first was collected in the lab and the second was collected on-line. We then use this information to determine how sample sizes must be adjusted when the experiment is conducted on-line.

On-line and lab settings differ in many respects, with likely consequences on effect sizes and variability. Whereas it is possible to select participants within a certain age range or socio-educational background in on-line participant pools, internet-based studies typically sample from a more diverse population than lab experiments. As mentioned already, this is an attractive feature of on-line studies. The observation that the majority of experimental results are collected with college students and the concern that this may hinder the external validity of the findings is known as the "college sophomore problem" and has been discussed in many fields including social psychology (Sears, 1986), marketing and consumer research (Peterson & Merunka, 2014), and political sciences (Druckman & Kam, 2011). The issue of whether psycholinguistic results generalize to other populations has been raised as well (Speed, Wnuk, & Majid, 2017). On-line and lab participants might apply different strategies. On-line participants might be less likely to discover the experimental manipulation or they might be less sensitive to the properties of the experimental material or the task, because they have no training in experimental psychology or linguistics. The two populations could also differ in their ability to perform meta-linguistic tasks (e.g., judge whether a sequence of letters is a word or a non word, or whether a sentence is grammatical or agrammatical). These differences could lead to smaller (or possibly larger depending on the experimental manipulation) effect sizes in on-line data, which would in turn have consequences for the statistical power of the experiment in this setting.

There are also several reasons to expect more variability in on-line than in lab data. We know that the timing of responses in an experimental setting depends on the hardware and experimental software. Several studies have examined the timing performance of web browsers commonly used to run on-line experiments (e.g., Bridges, Pitiot, MacAskill, & Peirce, 2020; Chetverikov & Upravitelev, 2016; Reimers & Stewart, 2015). In a recent study, Bridges et al. (2020) compared the precision of several web browsers and operating systems. Whereas the timing of experimental events was found to be more precise in lab-based settings (i.e., less inter-trial variability) than that obtained with web browsers, timing in the latter was surprisingly precise as well. The authors concluded that "modern computers are capable of very precise timing in presenting audio and visual stimuli and in receiving responses". They further replicated the finding that web browsers tend to present stimuli with longer gaps (between 8 and 40ms depending on the web browser/operating system combination) but that these gaps are constant in any browser/operating system combination. Information on the precision that web browsers can achieve and on potential differences between them is useful but does not tell the whole picture. In a real experiment, the experimenter has little control on technical aspects or on how these might differ across participants. Variation in timing from trial to trial is further influenced by the number of processes taking place on a given machine at a given point in time. When the experiment is run in the lab, hardware and software are kept constant across participants, and trial to trial variation is kept minimal by closing all other applications and preventing other processes, e.g. updates, from running. This is not possible in an on-line experiment. Moreover, additional variability is possible from the participants themselves. They are more likely to be distracted, and are likely to be less homogeneous in their behavior than a cohort of first year psychology students (provided that the on-line study targets a wider audience). The extent to which each of the potential causes of additional variability truly increases the variance in the data and how this in turn impacts the power of on-line studies is not well understood. Crucially, not all sources of variability will contribute to the power of the design to the same extent in all designs. For instance, if participants are more variable from trial to trial in an on-line experiment, this will mean that power will be lower -and that sample sizes will have to be increased- irrespective of the type of design. If data collected on-line show more variability in average response times across participants, larger sample sizes will be needed for studies requiring comparisons across groups of participants. Finally, if participants differ more from one another in the impact of the experimental manipulation in the on-line than in the lab setting, larger sample sizes will be needed for both within and between-participant designs. In order to assess the impact of the additional variability that on-line datasets presumably entail, it is therefore crucial to determine the source(s) of this variability.

Several studies have been conducted to determine whether the internet could be used to collect psycholinguistic data. Most of these studies are either direct replications of an existing lab experiment or a novel experiment with the same properties as an existing experiment. In several experiments, psycholinguistic effects could be replicated on-line (Angele, Baciero, Gómez, & Perea, 2023; Corley & Scheepers, 2002; Fairs & Strijkers, 2021; Vogt, Hauber, Kuhlen, & Abdel Rahman, 2021; but see Demberg, 2013; Enochson & Culbertson, 2015; Vogt et al., 2021, Experiment 1). Lab and on-line data are however rarely directly compared. As a consequence, existing studies provide little information on whether on-line data truly differ from lab data and if so, in which aspects. Most importantly, the criterion to assess the feasibility of an on-line experiment is usually whether the experimental manipulation produces a significant effect, as determined by the *p*-value. The underlying reasoning is that if an effect is significant, we can conclude that the data are not too noisy. Focusing on statistical significance can be misleading especially when statistical power is low. The *p*-value depends on the estimated effect size as well as on the uncertainty (standard error) of that estimate. When these estimates are noisy (uncertain), the *p*-value provides little information on the reliability of an effect (on whether this effect truly exists) or on the replicability of this effect (i.e., how easily a significant effect can be reproduced if the study is run again, see for instance Vasishth et al., 2018). This approach therefore makes it difficult to compare findings across studies, to gain knowledge on the common properties of on-line data, and on how exactly these might differ from that of lab data.

In the present study, we focus on data in which the measurement of interest is the time taken by the participant to initiate a vocal response following the presentation of a stimulus. The vocal responses of participants are recorded during the experiment and the dependent variable is extracted by measuring the time interval between the onset of picture presentation and the onset of the vocal response (i.e., naming or speech onset latency). Unlike response times obtained by button press, data from language production studies are not ready to be used once the experiment has been completed by the participant. Each vocal response must be checked for accuracy, and the onset of the vocal response must be set manually. Multiple studies have shown that voice keys are less accurate than manual alignments and more prone to errors (Duyck et al., 2008; Kessler, Treiman, & Mullennix, 2002; Rastle & Davis, 2002). This is most likely even more of an issue in an on-line setting, because the voice key cannot be adjusted for each participant. It is therefore particularly important to determine the sample sizes required to run this kind of study. If it turns out that on-line studies require many more participants than lab studies to achieve the same statistical power, data pre-processing might be too demanding to make on-line data collection a viable alternative to lab experiments.

At least two studies have recently examined on-line production data. Vogt et al. (2021) replicated the semantic interference effect in picture naming (longer naming latencies when the picture is presented with a written word of the same semantic category than when it is presented with an unrelated word, e.g., Lupker, 1979; Piai, Roelofs, & Schriefers, 2011; Schriefers, Meyer, & Levelt, 1990) in two of three on-line naming tasks. Fairs and Strijkers (2021) replicated a study they had previously conducted in the lab. Assuming that the data would be more variable on-line, they increased the number of participants. Participants were asked to name pictures whose corresponding names were either of high or low frequency. The frequency effect (shorter latencies for more frequent words) was replicated on-line. The authors did not assess the variability of their on-line data but plots of the distributions of the naming latencies indeed suggest a higher variability in on-line data. Fairs and Strijkers (2021) also compared effect sizes across settings and did not find any statistical difference. Both studies concluded that on-line data collection for language production experiments is feasible and provided guidelines on how to improve the quality of these data. These guidelines include advice on the selection of the web experiment builder, on participant recruitment and selection, on assessing the reliability of the technical equipment, or on assessing the participants' motivation.

Building on these studies, the present study asks whether effect sizes and variability differ across settings and examine the consequences of these differences. Whereas the conclusions will be limited to on-line data from production experiments, the set of analyses we present can be useful to compare the output of other response time experiments across settings. The first analysis compares effect sizes between lab and on-line data. The second analysis compares the overall variability across settings. The next three analyses examine three sources of variability: within-participant variability (i.e., are participants less consistent in their response times when performing an experiment on-line than when performing an experiment in the lab?), variability in average response times across participants (are participants more variable in their average response times in on-line settings?), and variability in effect sizes across participants (are participants more variable in the impact of the manipulation in on-line settings?). Power simulations are then used to determine how sample sizes for items and participants need to be adjusted when the experiment is performed on-line as opposed to in the lab.

## Experiments

The first dataset was collected in the lab and is published in Bürki and Madec (2022). The paradigm is a classical picture-word interference task. Participants saw pictures of objects with a superimposed distractor word. Their task was to name the picture out loud, ignoring the distractor. In some trials, distractor and target words were of the same semantic category. In other trials, they overlapped in their first phonemes. In the remaining trials, the two words were unrelated. Previous studies have shown that semantically related distractors lead to slower speech onset latencies than unrelated distractors (Lupker, 1979; Rosinski, 1977; see also Vogt et al., 2021 for the demonstration that the effect can be replicated on-line) and that phonologically related distractors facilitate naming compared to unrelated distractors (Posnansky & Rayner, 1977; Rayner & Posnansky, 1978 for early demonstrations of this phonological facilitation effect). Both effects have been replicated numerous times in the lab and were replicated in Bürki and Madec (2022). The second dataset was collected on-line for the present study with the sole purpose of replicating the lab experiment.

### Participants

For the lab study, forty-five native German speakers were tested. They were aged between 18 and 30 (Mean = 23.2, SD = 3.5) and did not report any hearing, psychiatric, or linguistic disorders. They were paid or given course credit for their participation. They were all students at the University of Potsdam. For the on-line study, participants were recruited on the on-line research platform Prolific (www.prolific.co). They were aged between 18 and 30 (mean age for these participants was 24.5 (SD = 3.5) and did not report any hearing, psychiatric, or linguistic disorders. These participants were paid for their participation. Data collection was interrupted as soon as 45 participants had completed the study (15 other participants started the study but their recordings were not uploaded properly on the server). The study received ethics approval from the ethics committee of the University of Potsdam.

**Materials**

Ninety German nouns (target words) were selected for the test trials. Each had a corresponding picture in the Multipic database (Duñabeitia et al., 2018). In addition, 180 German nouns were selected to be used as distractors. Target words and distractors were combined to create 360 stimuli in four conditions. In the semantically-related condition, each of the 90 pictures was associated with a distractor from the same semantic category (e.g., animals, vegetables, pieces of furniture). In the semantically-unrelated condition, the 90 distractors of the semantically-related condition were re-assigned to different pictures such that target and distractor words had no phonological or semantic relationship. In the phonologically-related condition, each of the 90 pictures was associated with a distractor word sharing the same onset (one to four) phonemes. In the phonologically-unrelated condition, the 90 distractors of the phonologically-related condition were re-assigned to different pictures such that target and distractor words had no phonological or semantic relationship. An additional 90 stimuli were created by combining the pictures with a line of Xs. These trials were not considered in the analysis reported in the present study. Eight additional pictures and 16 distractor words were selected and combined to be used as fillers or training items.

**Procedure**

The experiment started with a familiarization phase. The pictures were presented one by one on the screen together with their corresponding written name but no distractor. During the picture-word interference task, picture-word stimuli were presented one by one on the screen. Participants were asked to name each picture as quickly and accurately as possible. The stimuli were divided into five blocks, each featuring the 90 pictures but associated with different distractors. In both experiments, the randomization of the stimuli was programmed such that each target word appeared once and in a different condition in each block, and such that each experimental block had the same number of trials from each condition. In both experiments, trials had the following structure: a fixation cross was first displayed at the center of the screen, replaced after between 2200 and 2300ms by the picture-distractor stimulus. Vocal responses were recorded during the first 3000ms after the onset of picture presentation. Vocal responses were checked for accuracy and the speech onset latencies were set manually. The only difference between the on-line and lab experiments was that in the latter, the EEG signal was monitored during the task.

**Analyses**

**Effect sizes.** In a first set of analyses we compared effect sizes across settings. We analysed the phonological and semantic manipulations in separate models. For each manipulation, we implemented a mixed-effects model with setting, experimental manipulation, and their interaction as fixed- effects. Contrasts were set such that lab data had a value of -0.5 and on-line data of 0.5. This way, the intercept represents the grand mean and a positive estimate reflects longer naming latencies for on-line data. For both manipulations, unrelated trials had a value of -0.5 and related trials a value of 0.5. This way a positive estimate reflects longer naming latencies for related trials. The random part of the models included by-participant and by-item intercepts, by-participant and by-item random slopes for the main effects and interactions, and correlations between random intercepts and slopes. The models were run in the Bayesian framework, using the R package brms (Bürkner, 2017). The output of a Bayesian model is a posterior distribution. The posterior distribution provides information on the possible values of the parameters of interest and their probabilities. From the posterior distribution we can derive the point estimate and its 95% credible interval. The 95% credible interval is the interval in which we can be 95% certain that the true value of the parameter lies (e.g., Gelman & Carlin, 2014). It provides information about the uncertainty of the estimated parameter. Posterior distributions are a combination of priors and data. The priors were set as follows. For the intercept, $\mathcal{N}(1000,400)$, reflects the assumption that the value for the intercept lies with 95% probability between 200 and 1800ms, with a higher probability for values around 1000ms. For the slopes we used $\mathcal{N}(0,100)$ for the effects of the experimental manipulations and the interactions and $\mathcal{N}(0,200)$ for the effect of setting. The first reflects the assumption that the effect lies with 95% probability between -200 and 200ms with a higher probability for values around 0. The second that the effect of setting lies with 95% probability between -400 and 400ms with a higher probability for values around 0. We used truncated normal distributions for standard deviations, namely $\mathcal{N}_+(0,100)$ for all random terms and $\mathcal{N}_+(0,200)$ for the residual standard deviation. These priors are weakly informative priors, that is, they allow a wide range of possible values. Weakly informative priors ensure that the outcome is not constrained by the priors but mostly informed by the data. We also report Bayes factors for the interactions between setting and experimental effects. Bayes factors were computed with more informed priors for the interaction, namely, $\mathcal{N}(0,10)$, $\mathcal{N}(0,20)$, and $\mathcal{N}(0,40)$. Bayes factors are highly sensitive to the choice of priors. Priors that are too diffuse (i.e, uninformative) tend to favor the simpler model (see for instance Schad, Betancourt, & Vasishth, 2021).

**Overall variability.**

In a second set of analyses, we compared the variability in the dependent measure across settings (lab vs. on-line) using Bayesian distributional models. With these models, we can estimate the effect of a predictor on both the mean and the residual standard error. Our models did not include any random effect. Therefore, this analysis tells us whether the data are more variable in a given setting or in the other, but does not provide information on the source of this variability. For this analysis, we used the following weakly informative priors. For the intercept, $\mathcal{N}(1000,400)$ reflects the assumption that the value for the intercept lies with 95% probability between 200 and 1800ms, with a higher probability for values around 1000ms. For the slope, we used $\mathcal{N}(0,300)$; this assumes that the effect size lies with 95% probability between -600 and 600ms with a higher probability for values around 0. We used truncated normal distributions (only positive values are allowed) for sigma, namely $\mathcal{N}_+(0,50)$. With this analysis we can test the widely held assumption that data collected on the internet are more variable.

**Within-participant variability across trials.** With the next analysis, we asked whether a given participant tends to be consistent in their response times and whether this reliability, or consistency, depends on whether the data are collected on-line or in the lab. The following analysis was conducted separately for each dataset. The data of each participant was split in odd and even trials. For each participant, we computed the mean for each trial category. We then computed the correlation between the two means. Finally, we compared the correlations across settings (see Parsons, Kruijt, & Fox, 2019; Urbina, 2014 for similar procedures).

Naming latencies are influenced by item-specific properties and experimental conditions. As a result, a perfect correlation cannot be expected, even if participants are highly consistent in their response times. We are, however, not interested in the absolute value of the correlation but in differences in the strength of this correlation between lab and online data. A lower correlation for instance in the on-line experiment would suggest that the data are more variable in this setting, with consequences for all types of designs that involve repeated measures within participants (i.e., most designs in psycholinguistic research).

**Between-participant variability in response times and effect sizes.** In psycholinguistic studies, most designs involve repeated measurements. A given participant usually provides responses on multiple items (see Laurinavichyute & Malsburg, 2022, for an example of study using one trial per participant) and a given item is usually seen by multiple participants. Random factors in these studies are often crossed, i.e., the random factor *item* occurs at more than one level of the random factor *participant*. Mixed-effects models are one of the best approaches to analyze data from crossed designs (see Baayen, Davidson, & Bates, 2008; Matuschek, Kliegl, Vasishth, Baayen, & Bates, 2017). They are appropriate because they allow modeling the within-participant and within-item dependencies in the data in a single statistical model. As such, they ensure that the findings generalize to the populations of items and participants the samples were taken from. These models are often used for the sole purpose of testing for significance. In addition to providing a *p*-value, they also provide estimates of variability. In statistical terms, this information is contained in the variance components of the statistical model. The standard deviation of the by-participant random intercept is an estimation of variability in mean response times across participants and the standard deviation of the by-participant random slope is an estimation of variability in effect sizes across participants. Together with the residual standard deviation, the standard deviations associated with the random terms form the variance components of the model. Thus, in a given analysis, these standard deviations provide information on the different sources of variability in the data.

We first compared the standard deviation of the by-participant random intercept across settings for each experimental effect. These standard deviations estimate between-participant variability in response times. In linear mixed-effects models, the estimates for individual participants are shrunk towards the grand mean (Bates, Mächler, Bolker, & Walker, 2015). The less data we have for a participant or the more extreme the data compared to that of other participants, the greater the shrinkage, also called partial pooling (Gelman & Hill, 2007). For each manipulation, we quantified the standard deviation of the by-participant random intercept for each setting. We then compared these standard deviations using F-tests. If participant mean naming latencies differ more on-line than in the lab, larger sample sizes will be required in this setting for between-participant designs.

Then, we examined the extent to which participants vary in their sensitivity to the experimental manipulations by looking, for each experimental effect, at the standard deviation of the by-participant random slopes. Many psycholinguistic studies use within-participant designs, i.e., the same participants are tested in different conditions. In these designs, between-participant variability in mean response times will have little impact on the standard error of the estimate of the fixed effect. Inter-individual differences in the effect of the experimental manipulation, by contrast, will influence this standard error (and, as a consequence, the power, i.e., the probability of finding the effect in subsequent studies).

Contrasts were set such that unrelated trials had a value of -0.5 and related trials a value of 0.5. The random part of the model included by-participant and by-item intercepts, by-participant and by-item random slopes for the effect of relatedness, and correlations between random intercepts and slopes. The priors in these analyses were the same as in the first analysis. For each analysis, we display the posterior distribution of the standard deviations. Appendix A displays the point estimates of the standard deviations and their 95% credible intervals. The variances are compared across settings for each manipulation using F-tests.

**Power computations.** In the last series of analyses, we computed power for the two effects and settings. Power computations for mixed-effects model rely on simulations. Simulations in turn require estimates of effect sizes, estimations of between-item and between-participant variability, as well as of residual variability. For our simulations, we used the estimates of the statistical models described in the previous section. Simulations were run separately for each setting and experimental effect. We simulated power as a function of the number of participants, from 12 to 96 in steps of 12. We did so for 90 items (number of items in the original experiment) and repeated the procedure for 40 items and 20 items.

The data, materials, and code are available at https://osf.io/hbz9y/

## Results

**Accuracy**. The distribution of errors was similar in both settings. In the lab data, number of errors per participant ranged between 0 and 63, with a mean of 22 (6%). 15228 data points were used for the analysis. In the on-line data, number of errors per participant ranged between 1 and 91, with a mean of 26 (7%). 15024 data points were included in the analysis. The low number of errors in the on-line experiment suggests that our participants performed the experiment in good faith. Previous reports suggest that this is not always the case (e.g., Fairs & Strijkers, 2021) or that participants recruited on on-line platforms tend to make more errors (e.g., Vogt et al., 2021). We note that picture naming is a simple task, which does not require any metalinguistic judgment. This may explain why our on-line participants performed just as well as the lab participants. We further note that in the on-line experiment, the data of 15 participants were not recorded. For all the other participants, the sound quality was good enough and the data of all these participants could be included in the analysis.

**Effect sizes.**

Table 1: *Model comparing the size of the semantic interference effect across settings, estimates with 95% credible intervals*

| term | estimate | lower | upper |
| --- | --- | --- | --- |
| Intercept | 1,068.95 | 1,036.15 | 1,100.36 |
| Relatedness | 47.18 | 31.88 | 62.30 |
| Setting | 275.09 | 217.50 | 333.26 |
| Setting * Relatedness | -24.53 | -43.99 | -5.21 |

Table 2: *Model comparing the size of the phonological facilitation effect across settings, estimates with 95% credible intervals*

| term | estimate | lower | upper |
| --- | --- | --- | --- |
| Intercept | 1,024.49 | 992.98 | 1,056.49 |
| Relatedness | -26.11 | -42.70 | -9.49 |
| Setting | 280.29 | 224.80 | 334.56 |
| Setting * Relatedness | 9.92 | -9.47 | 28.91 |

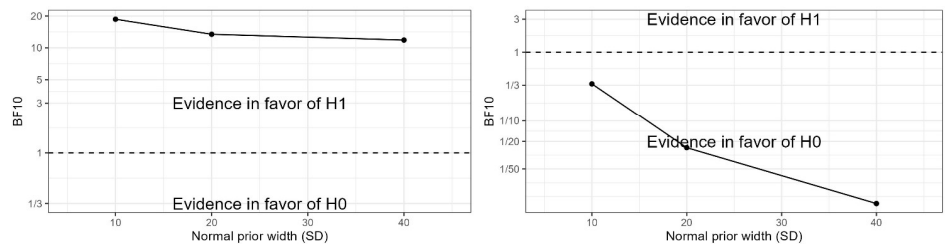

Figure 1: Bayes Factors for the interactions between setting and semantic effect (left) and between setting and phonological effect (right)

Tables 1 and 2 display the output of the models comparing the effects of the semantic and phonological manipulations across settings. The estimates for the interactions between setting and each of the experimental manipulations suggest that both effect sizes are smaller in the on-line than in the lab data. The results of Bayes Factor analyses for the interactions are displayed in Figure 1. For the semantic manipulation, they provide support for the hypothesis that the effect is smaller in on-line data than in the lab data. For the phonological manipulation, the evidence is inconclusive for the smallest prior and favors the null hypothesis of no effect with larger priors. We note that irrespective of whether the estimated differences in effect sizes across settings are supported by these analyses, if these estimates are used in power simulations, everything else being equal, these simulations will most likely show that power is lower in an on-line setting than in the lab.

**Overall variability.**

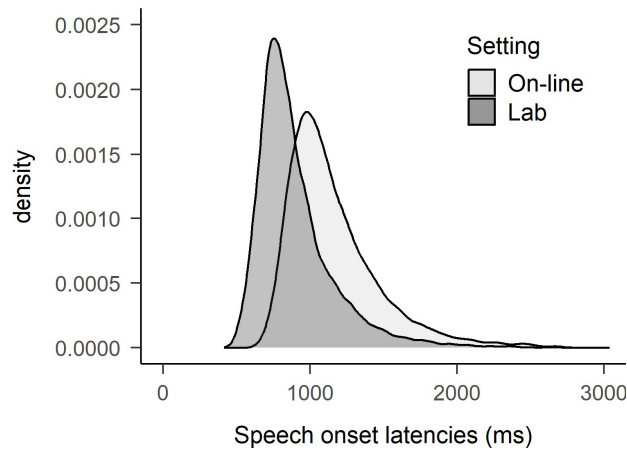

Figure 2: Density of speech onset latencies in on-line and lab data

Figure 2 displays the densities of the speech onset latencies for lab and on-line data. The distributional model comparing means and variances across settings shows that naming times are longer in on-line than in lab data (estimate of difference = 280ms [274-287]). The residual variance is also larger for on-line than for lab data (estimate of difference = 46 [42-50]). Subsequent analyses examine the source(s) of this difference.

**Within-participant variability.** Correlations between odd and even trials for the lab and on-line data are displayed in Table 3. The reliability of on-line and lab data is highly similar. Both correlations are high and do not differ from one another ($p = 0.22$). This analysis does not provide support for the hypothesis that response times fluctuate more for a given participant in an on-line than in a lab study.

Table 3: *Within-participant variability: Correlations between odd and even trials for each dataset*

| Setting | r | lower | upper |
|---------|------|-------|-------|
| Lab | 0.96 | 0.93 | 0.98 |
| On-line | 0.98 | 0.97 | 0.99 |

**Between-participant variability in response times and effect sizes**

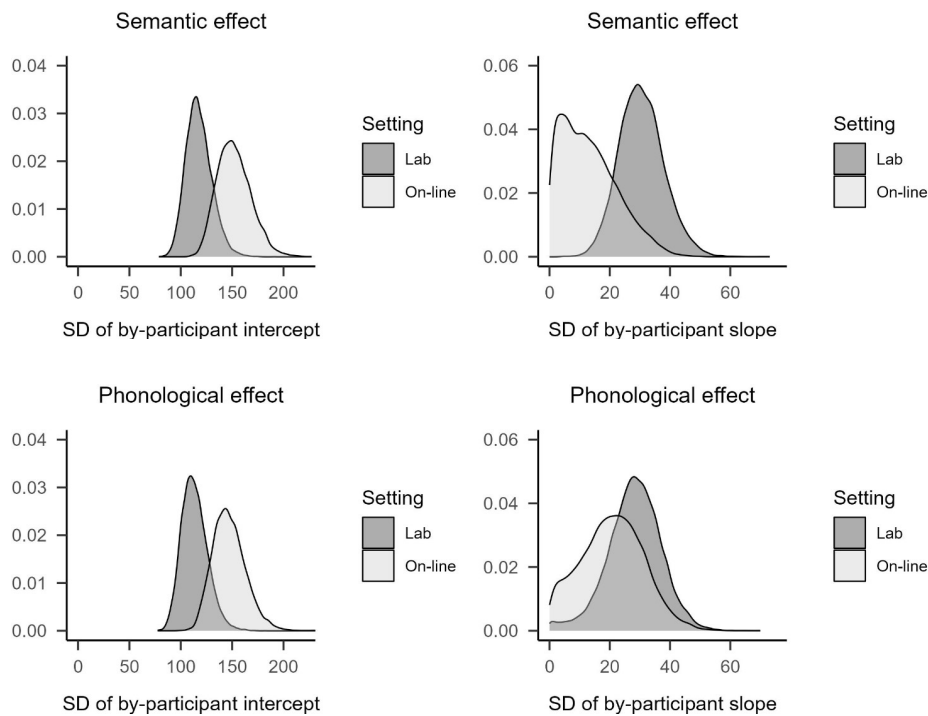

Figure 3: Posterior distributions (with 95% CrI) of the standard deviations of by-participant intercepts and slopes

The posterior distributions of the standard deviations of the by-participant intercepts and slopes are displayed in Figure 3. Descriptively, the standard deviations of the by-participant intercepts in the two models are higher in the on-line than in the lab data. The ratio of the two variances for the semantic effect is 1.67 and is significantly different from 1 ($p = 0.05$). The ratio of the two variances for the phonological effect is 1.68 and is significantly different from 1 ($p =0.04$). If the values of the respective models are used to compute the sample sizes required for a between-participant manipulation, the required sample sizes will be higher for the on-line data than for the lab data. This is because in such designs, the standard error of the fixed effect will tend to increase when the by-participant intercept increases.

For both effects, the standard deviation of the by-participant random slope has a slightly smaller estimate in the on-line than in the lab data. The standard deviation of the random slope for the phonological effect is 20.7 [1.6, 41.0] in the on-line data and 28.1 [8.1, 45.1] in the lab data. The standard deviation of the random slope for the semantic effect is 13.2 [0.6, 34.1] in the on-line data and 30.5 [16.8, 46.2] in the lab data. The ratio of the two variances for the semantic effect is 0.19 and is not significantly different from 1 ($p = 1.00$). The ratio of the two variances for the phonological effect is 0.54 and is not significantly different from 1 ($p =0.98$).

As can be seen on Figure 3 the posterior distributions of the by-participant slopes suggest that estimates are less precise for on-line data. Many values are at or around zero. This could suggest that there is not enough power to estimate the by-participant variability in effect sizes in this setting. To get a sense of the descriptive variability in effect sizes across participants, we computed the descriptive mean difference between related and unrelated conditions (separately for the semantic and phonological manipulations) for each participant. We then computed the grand average and standard deviation of this grand average. The standard deviations for the semantic manipulation are respectively 47.1 and 45.1 for the lab and on-line data. The standard deviations for the phonological manipulation are respectively 45.1 and 46.6. The values obtained this way are higher than that estimated by the model. This could suggest that the small values estimated by the models for on-line data do not reflect a lack of variability in effect sizes across participants but rather a lack of sufficient data to estimate the standard deviation of the by-participant random slopes. Unfortunately, this analysis does not allow us to conclude whether participants are more or less impacted by the experimental manipulation in on-line than in lab settings.

**Power computations**

An important issue in assessing the feasibility of on-line experiments concerns the sample sizes that these experiments require to achieve sufficient power. As discussed already, power depends on the (unknown) true effect size, the sample sizes, and the variance components. Our analyses show that both effects are descriptively smaller on-line. Power can only be computed with reference to a specific effect size. For the on-line data at hand, our best estimates are the estimates from the statistical models. Given that these estimates are smaller than that of the models for the lab data, we can expect lower power for the on-line setting. Power further depends on the variability of the data, but as already mentioned and explained at length in Westfall, Kenny, and Judd (2014), not all sources of variability will contribute to the power of the design to the same extent in a given design. In a within-participant design like the one used here, three sources of variability will impact power, the variability in effect sizes across participants (estimated by the standard deviation of the by-participant random slope), the variability in effect sizes across items (estimated by the standard deviation of the by-item random slope), and the residual error.

For both experimental manipulations in the present study, the standard deviation of the by-participant random slope is smaller in the on-line data than in the lab data and the residual variance is higher in this setting. A smaller standard deviation for the by-participant random slope should benefit power; by contrast, a higher residual variability will decrease the power. Results of power simulations for 20, 40 and 90 items are displayed in Figure 4. For the semantic manipulation, we can see that power is higher in the lab, and above 80% with 45 participants (number of participants included in the study) with 20, 40 and 90 items. In the on-line setting, power is above 80% with 90 items and 45 participants but requires at least 55-60 participants to reach 80% with 40 items. With 20 items, power remains below 60% even with 96 participants. For the phonological manipulation in the lab, power reaches 80% with 90 items and about 25 participants but it never reaches 80% with 40 or 20 items. On-line, power seems to reach 80% with 90 items and about 80 participants, it remains well below 80% with 20 and 40 items with 96 participants.

Our simulations further suggest that increasing the number of items often has more impact than increasing the number of participants. This observation echoes back to the simulations reported in Westfall et al. (2014), who showed that power benefits less from increasing the sample size of a random factor when the variability of this random factor is low. In the present data, the estimated standard deviations of the by-participant random slopes are lower than the estimated standard deviations of the by-item random slopes. The same pattern (i.e., increasing the number of items has more impact than increasing the number of items) was reported in Vogt et al. (2021), in which the standard deviation of the by-participant random slope was set to 0. Moreover, beyond a certain number of participants, and especially with a low number of items, power does not seem to increase much with the addition of new participants. This is again in line with the simulations of Westfall et al. (2014), who showed that in designs with few items, power may converge to a value which is well below 80%.

In order to estimate the impact of the amount of unexplained variance on statistical power, we also simulated power as a function of residual standard deviation (from 100 to 300 in steps of 50) for the phonological effect in the on-line dataset. The results of this simulation can be seen in Figure 5. They show that a difference of about 50 (which corresponds to the difference we observe between on-line and lab data) decreases power by a little less than 10%.

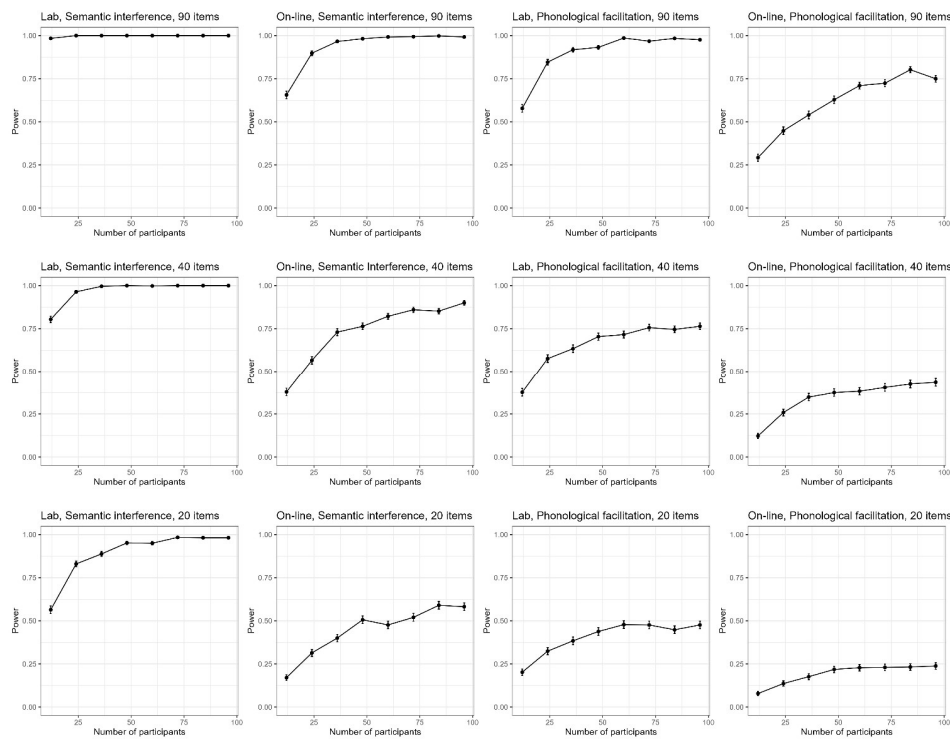

Figure 4: Power to detect the semantic interference and phonological facilitation effects as a function of number of participants for 90 items, 40, and 20 items in lab and on-line data

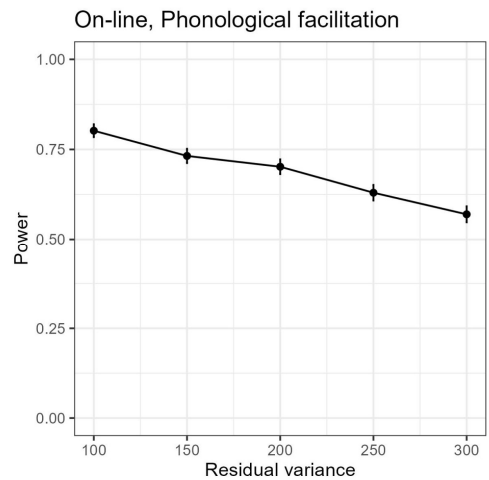

Figure 5: Power to detect the phonological facilitation effect in an on-line setting as a function of residual variance (with 90 items and 45 participants)

## General Discussion

The use of the internet for experimental data collection is becoming increasingly popular. Yet, for most experimental effects, the expected effect sizes in on-line settings, the amount of variability in these settings, and the sources of this variability are not known. Without this information, it is difficult to estimate the sample sizes required to design studies with sufficient statistical power. We compared effect sizes and variability across settings for the same picture naming experiment and used this information to determine how sample sizes have to be adjusted when the experiment is conducted on-line.

Comparisons of effect sizes across settings show that the two effects examined in the present study are descriptively smaller on-line than in the lab. The semantic interference effect (longer response times to name pictures with semantically related than with unrelated distractors or semantic interference effect) decreases by 25ms in the on-line setting. The phonological facilitation effect (shorter response times to name pictures with phonologically related distractors than with unrelated distractors or phonological facilitation effect) decreases by -10ms in the on-line setting. Bayes factors provide support for the hypothesis that the semantic interference is smaller in the on-line than in the lab data. A likely explanation of this finding is that the true effect size depends on the population being tested. Whereas the sample for the lab experiment mostly consisted of undergraduate students with backgrounds in linguistics or psychology, the background of participants in the on-line study was more diverse. The two populations might differ in the strategies they used to perform the task, in the extent to which they noticed the similarities between distractors and target words, in their ability to ignore the distractor, or in the relative speed with which they processed the two words (Bürki et al., 2019; Bürki & Madec, 2022). It is important to note, however, that this finding cannot be generalized to other paradigms or effects without further investigation. It is possible that true effect sizes for other manipulations are more similar in the two settings or even larger in on-line settings. The observed difference in effect sizes across settings for the semantic manipulation in the present study shows that effects cannot simply be assumed to be of the same size in on-line and lab experiments. Power depends on effect size (inter alia). If the true effect size of an experimental effect is smaller on-line than in the lab, the statistical power of the design will be lower when the study is conducted on-line, irrespective of whether the design involves comparisons within or between participants.

Data collected on-line are often assumed to be more variable than data collected in the lab. In order to compensate for this variability, researchers often increase their sample sizes, or advise others to do so (e.g., Vogt et al., 2021). Importantly, however, whether the sample size needs to be increased for a given design will depend on the source(s) of the variability. In a within-participant design, there are at least four sources of variability: variability within participants across trials, variability between participants in average response times, variability between participants in effect sizes, and residual variability. Our analyses first show that within-participant across trial variability was very low in both settings, with no statistical difference between the two settings. It is sometimes argued that participants are more distracted at home than in the lab or that the timing of experimental events fluctuates more during an on-line experiment. These hypotheses predict a lower within-participant across trial consistency, with consequences for the statistical power of all types of designs. Our analyses do not confirm these predictions for the data at hand.

Our analyses further show that the standard deviations of the by-participant intercepts were significantly higher in the on-line than in the lab data, as predicted under the hypothesis that participants vary more in their overall response times in on-line studies. It can be hypothesized that at least part of this additional variability comes from the greater diversity of the participants in the on-line sample. The difference in the standard deviations of the by-participant random intercepts between the two settings is greater than the lags that are often observed across combinations of operating systems and web browsers in studies without real participants (e.g., Bridges et al., 2020) suggesting that technical differences between lab and on-line settings are not sufficient to explain the greater between participant variability in the on-line setting. Several authors have suggested that on-line data collection might be less suitable for between-participant designs. The present analysis provides statistical support for this suggestion. If such studies are conducted on-line, larger sample sizes will likely be necessary to achieve the same statistical power than in the lab. Fortunately, the majority of studies in psycholinguistics (as well as in many other subfields of experimental psychology) focus on effects that can be manipulated within participants. For these studies, the power of the experiment will not be affected by differences in response times but will mostly depend on the amount of variability in effect sizes across participants (Westfall et al., 2014).

An estimation of the variability in effect sizes across participants is provided by the standard deviation of the by-participant random slopes. In the data analyzed here, estimates of this component were unexpectedly low. They were much lower than the standard deviations of the by-item random slopes, and much lower than the descriptive standard deviations of the between participant differences in effect sizes, for both, lab and on-line data. Notably, we also observed that the standard deviation of the by-participant random slope was often estimated with lower precision in the on-line data, with many values around and at zero (see Figure 3). Such patterns are common when the statistical model does not have enough data to estimate a random term. In the frequentist framework, the model often returns a convergence warning when this happens. In the Bayesian framework, the model can converge but settles on a value that is below the true value. It can therefore be hypothesized that the true values of this variance component are higher than estimated here, i.e., closer to the descriptive estimates. As a result of this imprecision in the estimation, comparisons of the standard deviations of the by-participant random slopes across settings are not informative (note however that descriptive estimates are highly similar across settings). Larger data sets would presumably be necessary to estimate this variance component with more precision.

Finally, our analyses show that residual variability is higher in the on-line than in the lab data, suggesting that on-line data contain a larger amount of variability that cannot be ascribed to the random terms. We note that this greater variability could at least partly be due to the greater uncertainty in the estimation of the by-participant random slope in the on-line data. Assuming that there is variability between participants that the model cannot associate with the by-participant random slope, this variability will end up in the unexplained residual variability. Importantly, residual variability being the largest source of variance, it will have the largest impact on the statistical power of any design compared to the other sources of variability. Given that residual variability is greater on-line, we can expect the power of the design to be lower in that setting for any given effect size and sample size.

As a result of differences in effect sizes and residual variability (between participant differences in overall naming times are not relevant for the power of our design), statistical power was expected to be lower on-line than in the lab. Our simulations confirmed this pattern. For the phonological facilitation effect and the original number of items (90), 25 participants are sufficient to obtain a power of 80% in the lab but at least 84 participants are necessary when the data are collected on-line. Notably, when the number of items is reduced to 40 (which is still more than in many studies in the field), power for the on-line data seems to converge to a value well below 80%. These findings have important practical consequences. First, assuming that the required number of participants to achieve sufficient power can be tested, the human resources and therefore the time and money required to pre-process the data will be much higher if the experiment is conducted on-line, i.e., more than three times higher in our example. In the experiment considered in the present study, each participant produced 90 items in the related and unrelated conditions. The 180 vocal responses for each participant have to be checked manually to detect errors and speech onsets, i.e.,180 * 25 =4,500 in the lab, and 80 * 84 =15,120 on-line. Second, experiments with a low number of items may never reach sufficient power, irrespective of the number of participants. We note that this is not specific to on-line data and has already been demonstrated elsewhere (Westfall et al., 2014). Our simulations show that this situation might occur more often for on-line than for lab experiments, at least with the current design. Increasing the number of participants no longer seems to impact power after a certain point because the data are simulated under the assumption that participants do not differ much in their effect sizes. If it is true that participants do not differ much in the impact of the experimental manipulation (and differ even less on-line than in the lab) then increasing power will be better achieved by increasing the number of items. As discussed above, there are however good reasons to assume that the actual variability in effect sizes across participants is greater than estimated by the models. If this hypothesis is correct, then power is overestimated in our simulations and the contribution of adding new participants is underestimated.

At least two previous studies estimated the number of participants needed to reach a power of 80% or higher for an on-line language production experiment, and gave recommendations with regard to the sample sizes required for such experiments. How do our power analyses compare to theirs? Vogt et al. (2021) used a picture-word interference paradigm and manipulated the semantic relationship between target and distractor. Their participants performed a naming task (as in the present study) as well as a categorization task with the same stimuli. Vogt et al. (2021) ran the study on-line only, using different web experiment builders and instructions. On the basis of power simulations, they concluded that an increase of 33% of the participant sample should be considered when running the experiment on-line. Our simulations show that this will not always be sufficient and that the required increase in participant sample size depends on the effect being tested as well as on the number of items. We further note that their simulations were ran under the assumption that the participants did not vary in their effect sizes with, as a likely consequence, an overestimation of the power of the design. Fairs and Strijkers (2021) asked participants to name pictures and manipulated, among other things, the frequency of the picture name. On the basis of simulations, they concluded that 40 participants might be sufficient for the type of experiments they conducted. This is clearly less than in the present study. We note however that in their study, each participant named over 200 items. Our simulations do not show whether 40 participants would be sufficient with 200 items, we suspect that it might be the case. In our simulations we considered numbers of items that are frequently used in psycholinguistic studies. With these numbers, 40 participants will rarely be sufficient in an on-line setting.

The fact that we observe differences across studies is not surprising. First, the effect sizes entered in these power simulations are point estimates. For instance, in the power computations for the semantic interference effect in the on-line study, we used the value of 34.37 because the model estimated that this was the most likely value. The credible interval for this effect shows that values between 16.8 and 51.8 are also plausible. As a result, different studies will most likely use different estimates. In addition, different experimental manipulations will be associated with different true effect sizes, with possible differences between on-line and lab settings. Finally, on-line studies might differ from one another in technical aspects as well as in procedural aspects and these differences may impact the quality of the data. The web experiment builder used to collect on-line data has for instance been shown to have an impact on the length of recordings as well as on whether an effect previously found in the lab could be replicated on-line. In Vogt et al. (2021), for instance, JsPsych performed better than SoSciSurvey. In the present study, we used PCIbex (Zehr & Florian, 2018), yet another platform. As discussed already, the high within-participant consistency and the high accuracy rates speak to the quality of the data. The extent to which our findings can be generalized to other web experiment builders remains to be examined. Notably, the fact that we observe differences across studies suggest that sample size recommendations should be given and considered with care.

**Conclusion**

The present study complements recent attempts to determine whether response times collected over the internet can be used to address fundamental research questions in the psycholinguistic study of language production. The comparison of lab and on-line data for the same experiment supports the hypothesis that data collected over the internet are more variable than data collected in the lab and shows that effect sizes may differ across settings. These differences between lab and on-line data have important consequences for the statistical power of the design. They show that sample sizes, even for within-participant designs, might have to be drastically increased in on-line settings to achieve sufficient power. This in turn reduces the attractiveness of on-line settings for experiments which require manual processing of the participant responses. More generally, our findings highlight the need to carefully assess differences in effect sizes and sources of variability across settings before moving a new paradigm on-line and provide a set of analyses that can be used for this purpose. These analyses can also be used to compare data collected with different web experiment builders or participant pools.

## Declarations

**Funding**

This research was funded by the Deutsche Forschungsgemeinschaft (DFG, German Research Foundation) – project number 317633480 – SFB 1287, Project B05 (Bürki) and project Q (Vasishth and Engbert).

**Conflicts of interest**

The authors have no relevant financial or non-financial interests to disclose

## Acknowledgments

The authors would like to thank the members of the Cognitive Science: Language and Methods Lab at the University of Potsdam for their useful comments, as well as Julia Pantelmann for her help in data collection and processing for the on-line study.

**(APPENDIX) Appendix**

**Output of Bayesian mixed-effects models for each setting and experimental manipulation**

Table 4:

*Output of mixed-effects model, lab data, semantic manipulation*

| term | estimate | conf.low | conf.high |
|---|---|---|---|
| Intercept | 930.84 | 892.68 | 970.55 |
| Relatedness | -59.59 | -77.57 | -41.25 |
| by-item random intercept | 77.87 | 66.68 | 91.96 |
| by-item random slope | 55.38 | 40.24 | 72.07 |
| by-participant random intercept | 116.60 | 95.87 | 144.65 |
| by-participant random slope | 30.20 | 16.80 | 46.16 |
| by-item correlation | -0.22 | -0.49 | 0.07 |
| by-participant correlation | -0.73 | -0.94 | -0.36 |
| Residual error | 236.92 | 233.10 | 240.82 |

*Output of mixed-effects model, lab data, phonological manipulation*

| term | estimate | conf.low | conf.high |
|---|---|---|---|
| Intercept | 884.01 | 848.72 | 920.41 |
| Relatedness | 30.85 | 11.94 | 49.76 |
| by-item random intercept | 76.07 | 65.21 | 90.12 |
| by-item random slope | 64.28 | 50.24 | 80.74 |
| by-participant random intercept | 112.42 | 91.79 | 140.93 |

| term | estimate | conf.low | conf.high |
|---|---|---|---|
| by-participant random slope | 28.43 | 8.11 | 45.09 |
| by-item correlation | -0.07 | -0.32 | 0.20 |
| by-participant correlation | -0.11 | -0.55 | 0.34 |
| Residual error | 223.56 | 220.01 | 227.24 |

*Output of mixed-effects model, on-line data, semantic manipulation*

| term | estimate | conf.low | conf.high |
|---|---|---|---|
| Intercept | 1,193.20 | 1,144.40 | 1,239.78 |
| Relatedness | 34.37 | 16.84 | 51.78 |
| by-item random intercept | 76.45 | 63.03 | 92.91 |
| by-item random slope | 53.11 | 34.85 | 71.86 |
| by-participant random intercept | 150.69 | 124.17 | 186.70 |
| by-participant random slope | 11.88 | 0.60 | 34.07 |
| by-item correlation | -0.44 | -0.67 | -0.13 |
| by-participant correlation | 0.02 | -0.69 | 0.74 |
| Residual error | 272.85 | 268.43 | 277.36 |

*Output of mixed-effects model, on-line data, phonological manipulation*

| term | estimate | conf.low | conf.high |
|---|---|---|---|
| Intercept | 1,177.61 | 1,130.76 | 1,221.87 |
| Relatedness | -21.50 | -40.11 | -2.47 |

| term | estimate | conf.low | conf.high |
| --- | --- | --- | --- |
| by-item random intercept | 74.64 | 61.95 | 90.29 |
| by-item random slope | 62.33 | 46.46 | 80.14 |
| by-participant random intercept | 145.71 | 119.86 | 182.65 |
| by-participant random slope | 20.84 | 1.61 | 40.98 |
| by-item correlation | -0.34 | -0.58 | -0.04 |
| by-participant correlation | 0.21 | -0.41 | 0.78 |
| Residual error | 264.69 | 260.43 | 269.03 |

**Power simulations using descriptive standard deviations of between-participant differences in effect sizes**

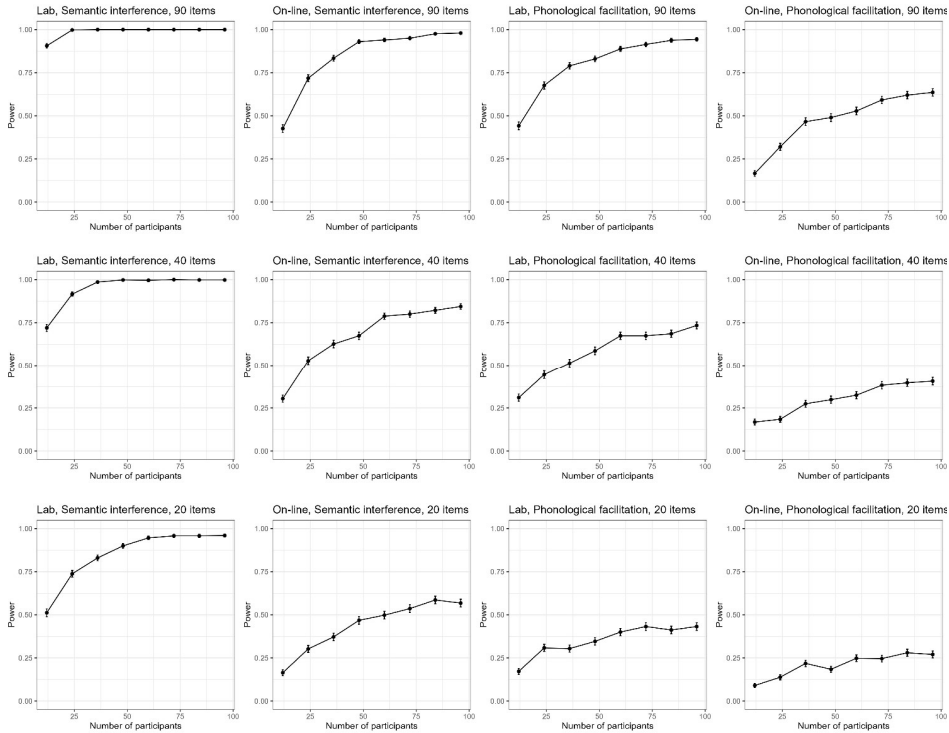

Figure 6: Power to detect the semantic interference and phonological facilitation effects as a function of number of participants for 90 items, 40, and 20 items in lab and on-line data

For each setting and experimental manipulation, simulations were based on the estimates from the corresponding Bayesian statistical models for all fixed effects and variance components except for the standard deviation of the by-participant random slope. For this parameter, we used the descriptive estimate. For each participant, we first computed the mean difference between the related and unrelated conditions. We then computed the grand average of this difference and its standard deviation. The value of the standard deviation was used for the simulations.